# FAKE OR GENUINE? CONTEXTUALISED TEXT REPRESENTATION FOR FAKE REVIEW DETECTION

Rami Mohawesh[1], Shuxiang Xu[1], Matthew Springer[1],
Muna Al-Hawawreh[2] and Sumbal Maqsood[1]

[1]School of Information and Communications Technology, University of
Tasmania, Tasmania, Australia
[2]School of Engineering and Information Technology, University of New South
Wales, Australian Defence Force Academy (ADFA), Canberra, Australia

## ABSTRACT

*Online reviews have a significant influence on customers' purchasing decisions for any products or services. However, fake reviews can mislead both consumers and companies. Several models have been developed to detect fake reviews using machine learning approaches. Many of these models have some limitations resulting in low accuracy in distinguishing between fake and genuine reviews. These models focused only on linguistic features to detect fake reviews and failed to capture the semantic meaning of the reviews. To deal with this, this paper proposes a new ensemble model that employs transformer architecture to discover the hidden patterns in a sequence of fake reviews and detect them precisely. The proposed approach combines three transformer models to improve the robustness of fake and genuine behaviour profiling and modelling to detect fake reviews. The experimental results using semi-real benchmark datasets showed the superiority of the proposed model over state-of-the-art models.*

## 1. INTRODUCTION

The Internet's size and importance has exploded in recent years, and it exerts a significant and growing influence on people's daily lives. Customers usually spend a substantial amount of time online, searching for information on a variety of products, communicating with others, and reading reviews. Additionally, the Internet enables individuals to write reviews on a range of topics based on their expertise and the opinions of others who have seen their work online. These individuals may use their reviews to promote or criticise various products or services [1, 2]. Consumers tend to purchase products that have a high number of good evaluations, which can lead to increased profits for the provider [3]. At the same time, negative reviews can result in financial losses for the companies involved [4]. Since anyone can write reviews without restriction, it is possible to provide undeserved positive or negative feedback with respect to products, services, and enterprises. Hence, it is necessary to verify the truthfulness of opinions and reviews posted online to assist people to avoid being misled by false information.

Ott, et al. [5] stated that human judges have been found to be quite poor at identifying fake reviews, with an accuracy of around 57%. Additionally, it is also difficult to conduct this type of identification manually. Detecting fake reviews using automatic detection based on state-of-the-art intelligent technologies has been found to be considerably faster and more accurate than using a human expert.

Many studies have been conducted into developing automated fake review detection models based on classic machine learning [6]. However, many of these models have some limitations resulting in low accuracy in distinguishing between fake and genuine reviews. These models have focused only on linguistic features to detect fake reviews and have failed to capture the semantic meaning of the reviews. Therefore, there is still a need for new models that are able to detect fake reviews efficiently. Recently, transformer or advanced pre-trained architectures have attracted considerable attention in text classification tasks and obtained superior results compared to the previous state of the art methods [7-9]. The effectiveness of these transformer architectures or techniques in constructing deep contextualized embeddings for a variety of texts is the key motivation to use them to develop a new model for detecting fake reviews. However, due the high variance and dynamic change of fake review features, using a single model could not provide a best fit for the entire training data. One transformer may be sensitive to the provided features and biased to specific features, leading to poor performance. Therefore, we present an ensemble approach that combines multiple transformer architectures. The ensemble learning has proven effective and achieved good results in text classification [10, 11]. The study reported in this Chapter was conducted to answer the following question: **Does the ensemble of transformer models perform better than state-of-the-art fake review detection methods?** As a consequence, we have proposed a new ensemble fake review detection model that combines three transformer models, namely, RoBERTa, ALBERT, and XLNet to improve the robustness of fake and genuine review profiling and modelling by handling the dependencies between input and output with attention and recurrence completely.

The main contributions of this paper are as follows:

- We investigate the performance of some transformer models in fake review detection. To the best of our knowledge, no previous study has used XLNet and ALBERT transformer models for fake reviews detection.
- We propose an ensemble model that combines three transformers called, RoBERTa, XLNet, and ALBERT to enhance the accuracy of fake review detection.
- The proposed model is compared with the state-of-the-art models and demonstrate superior performance. It significantly outperforms eight of the most recent fake review detection models.

The paper is organised as follows: Section 2 illustrates the related work done on this research topic. Section 3 provides a Preliminaries of the proposed model. Section 4 describes the proposed methodology in detail. Section 5 provides the experiments setting, the datasets and pre-processing, the evaluation metrics. Section 6 describes the results and discussion. Then, the paper is concluded in Section 7.

## 2. RELATED WORKS

Naive Bayes (NB) and Support Vector Machine (SVM), are examples of traditional machine learning algorithms that learn discriminant characteristics from reviews and have been used by a number of researchers to detect  fake reviews [12],[6],[13]. For example, using the Linguistic Inquiry and Word Count (LIWC) tool [14], Ott, et al. [5] proposed a model to automatically identify fake reviews by combining psychological features from reviews with n-grams, which were then fed to support vector machines. This approach achieved 90% accuracy in fake review categorisation, which is substantially better than human judges were able to achieve. According to the Op-Spam dataset, human judges were only 60% accurate. Feng, et al. [15] used a combination of context-free grammar and Part-of-speech features to detect fake reviews. Their results showed that these combined features could significantly increase fake review detection performance compared to the baseline method. Later, Li, et al. [16] developed a fake review detection model, titled Sparse Additive Generative Model (SAGE) which uses topic modelling [17] with a generalised additive model [18]. The proposed model results on Op-Spam and deception datasets achieved good results with accuracy of 81.82%, 83.10%, respectively.

Cagnina and Rosso [19] combined character n-grams, LIWC, and emotion features for fake reviews detection. Then, these extracted features are fed to an SVM algorithm to classify the reviews. Xu and Zhao [20] developed a model-based deep linguistic feature for fake reviews detection. Then they used an SVM classifier to classify reviews. Fusilier, et al. [21] proposed a model based on character n-grams content features to detect fake reviews. Then NB was used to detect fake reviews. Recently, several deep learning models such as recurrent neural network (RNN), convolutional neural network (CNN), and gated neural network (GRU) have been extensively used and achieved excellent results in the natural language processing field [22] and cybersecurity field [23], [24]. These models deal with dimensional data and extract the semantic presentation. Motivated by this, Ren and Zhang [25] introduced a recurrent convolutional neural network method with an attention mechanism to learn document representation. The proposed model proved its efficiency in detecting fake reviews. Similarly, and by using the context information of the sentences, Ren and Ji [26] also used deep neural networks and proposed a hybrid fake reviews detection model (GRNN–CNN). They combined a gated recurrent neural network (GRU) and a convolutional neural network (CNN). Their proposed model tested on the deception dataset and achieved good results with an accuracy of 83.34%. Later, Zhang, et al. [27] developed a recurrent convolutional deep neural networks model (DRI-RCNN) for fake reviews detection based on word contexts. Their proposed model performed well with 82.9% and 80.8% accuracy on spam and deception datasets, respectively.

More recently, Mohawesh, et al. [6] investigated some promising deep learning and transformers models. According to their experimental results, the RoBERTa transformer exceeded the performance of the state-of-the-art methods with a 91.02% accuracy on the deception dataset. They also found that BERT, DistilBERT, and RoBERTa performed very well with a small dataset. Although various machine learning technologies have been proposed to address fake reviews detection and to aid in distinguishing between fake reviews and genuine ones, it is rarely focused on contextualised text representation models. Thus, this work proposes a new ensemble model which combines three current states of the art deep learning models that can be used with any type of neural classifier and with any type of contextualised text representation and provides a comparative analysis of the performance of several pre-trained models and neural classifiers for fake review detection.

## 3. PRELIMINARY

Bidirectional Encoder Representations from Transformers (BERT): A BERT-trained model is used to pre-train a deep bidirectional representation of the text that can handle the unlabelled data by focusing on both right and left context in all layers simultaneously[28]. BERT was pre-trained on English Wikipedia text of about 2.5 billion words, consisting of an 800-million-word corpus of books. The BERT model reads a complete sequence of words in parallel, giving the model the ability to understand each word's context as a result of what is located around it. Fig 1 shows the BERT-base-cased model. It consists of 12 layers, 768 hidden layers, 12 heads and 109 million parameters. As shown in Fig 1, input is provided through a CLS token followed by a sequence of words. CLS is a categorisation token in this case. The input is subsequently passed to the layers above. Each encoder applies self-attention, transfers the result through a feed forward network, and then passes the result on to the next encoder in the sequence of layers. We obtained the output from the last transformer block, as shown in Fig 1.

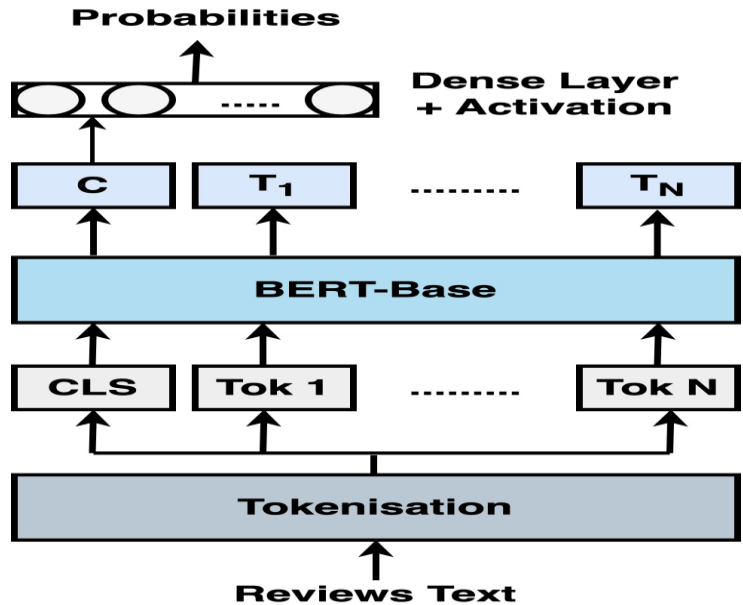

Figure 1. BERT model for text classification.

## 4. THE PROPOSED MODEL

As shown in Figure 2, the proposed model consists of pre-processing, combines the most recent transformer architectures and ensemble approach. The details of these stages are described as follows:

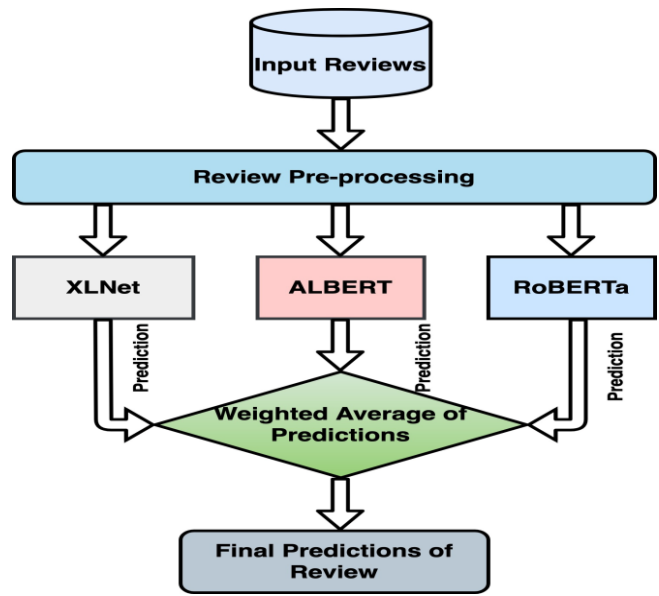

Figure 2. Ensemble of advanced pre-trained models' architecture.

## 4.1. Pre-Processing stage

To prepare the incoming reviews data as input for transformers, we perform many processes to clean the data and prepare it as input for transformers. We removed noise and irrelevant words, such as URLs and emojis, during this stage. Then, we split and divided each review text into a sequence of words.

**4.2. Transformers:** The proposed model combines the following three architectures:

**Robustly optimised BERT approach (RoBERTa):** an updated version of BERT that outperforms the BERT transformer [29] by training the model for a longer time on larger sequences and omitting the subsequent sentence prediction. Along with the English Wikipedia and book corpora, RoBERTa is pre-trained on the Common Crawl News dataset, comprising 63 million English-language news stories. RoBERTa base architecture consists of 768 hidden layers, 12 layers, 125 million parameters, and 12 attention heads. We used BoBERTa due to its high ability in providing dynamic masking patterns for each provided review.

**XLNet** [8]: It employs Transformer-XL [30] as a feature engineering model to gain a better understanding of the language context, which is an adaptation of the native Transformer. The Transformer XL model incorporates Relative Positional Encoding (RPE) and Recurrence Mechanism components into the Transformer used in BERT to manage long-term dependencies for texts that exceed the maximum permitted input length. An enormous dataset was used to train the XLNet model, which uses permutation language modelling. These permutations are one of the fundamental differences between BERT and XLNet, and they allow for the simultaneous generation of data from both sides. Thus, XLNet is able to learn the best features from the bidirectional review context which represent the fake review efficiently. The XLNet-base architecture consists of 768 hidden layers, 12 layers, 110 million parameters, and 12 attention heads.

**ALBERT:** is a pre-training natural language representation using modern language models that involves increasing the model size and number of parameters [9]. As a result of memory limitations and lengthier training hours, they can often be challenging to do. ALBERT (A Lite BERT) [9] employs parameter reduction strategies to boost model speed and reduce memory consumption to solve these difficulties. It shares the parameters among layers which provides a high-capacity contextual representation. This way of learning leads to capturing the meaning of words and improving the understanding of the entire text of the review. The A Lite BERT model achieved better results than DL models [31]. In our work, Albert-base-v2 architecture consists of 768 hidden layers, 12 layers, 12 attention heads, 128 embedding, and 11 million parameters.

We added a classification head with a single linear layer to each of the transformers mentioned above to distinguish between legitimate and fake reviews and label them. Then, the full architecture of each transformer and its parameters is fine-tuned to learn the review context.

## 4.3 Ensemble approach:

To gain the full benefits of transformers and their different perspectives on the learned features, we used the weighted average ensemble approach. In this approach, we extracted the output of the last layer of each transformer, which are logits. Then, we converted logits to probability using the SoftMax function. By using the weighted average of probabilities, we obtained a new probability for each class. The class with maximum probability is the predicted class. This approach gives the better transfer of the higher weight, improving the fake review detection rate and accuracy and preventing the bias of final decision to the less accurate model.

## 5. EXPERIMENT

This section presents the datasets descriptions, datasets pre-processing, the evaluation metrics and our model results and compares them to state-of-the-art approaches.

### 5.1. Experimental setup

We used a mini-batch size of 32 to train them on all of the datasets for 10 epochs. Overfitting was avoided by using an early stop [32]. While delta is set to zero, validation loss was used as the metric for early stopping [33]. We used AdamW optimiser [34]. Finally, we computed the loss using binary cross-entropy [35].

To evaluate the performance of the proposed model, we developed a python script and used Google CoLab interface[36]. We also used the transformer library [37], which was created by the *Huggingface* team. We run the experiments multiple times with different parameters. Table 1 presents the best hyperparameters of the three proposed models.

Table 1. The hyperparameter of the proposed models

| Models | Learning Rate | Batch Size | Type | Optimiser | Epochs | Max Length |
|--------|---------------|------------|------|-----------|--------|------------|
| RoBERTa | 2e-5 | 64 | roberta-base | AdamW | 15 | 256 |
| ALBERT | 2e-5 | 32 | albert-base-v2 | AdamW | 20 | 256 |
| XLNet | 2e-5 | 32 | xlnet-base-cased | AdamW | 20 | 256 |

### 5.2. Dataset Description

In this study, we used two publicly available benchmark datasets. **OpSpam** [5], and **Deception dataset** [16]. **OpSpam** dataset contains 1,600 review texts for twenty hotels in the Chicago area of the United States of America, 800 of which are fake and 800 of which are genuine. A label of '1' indicates fake reviews, whereas a label of '0' indicates legitimate reviews. These reviews came from a variety of sources. The fake reviews were constructed using Amazon Mechanical Turk (AMT), and the remaining were collected from various online review sites such as Yelp, TripAdvisor, and Expedia. **The the deception dataset** [16] represents a gold standard dataset containing 3,032 reviews. This dataset contains information about three distinct domains (hotels, doctors, and restaurants). Both datasets have only review text without any metadata information. In our experiments, 80% of the OpSpam and Deception datasets, were used for training, and the remaining 20% of each dataset was used to test the model. Table 2 shows The Statistical information for both datasets.

Table 2. The Statistical information of the OpSpam and Deception datasets.

| Datasets | Domain | Reviews Type | No. of reviews | No. of unique words | No. of sentences |
|----------|--------|--------------|----------------|---------------------|------------------|
| **OpSpam**[5] | Hotels reviews | Fake | 800 | 14427 | 7192 |
| | | legitimate | 800 | 14812 | 7963 |
| **Deception** [16] | Restaurant reviews | Fake | 201 | 5136 | 1827 |
| | | legitimate | 201 | 5126 | 1892 |
| | Doctor | Fake | 356 | 5128 | 2369 |

| | reviews | legitimate | 200 | 5098 | 1151 |
|---|---|---|---|---|---|
| | Hotel reviews | Fake | 1080 | 16,635 | 8463 |
| | | legitimate | 1080 | 17,328 | 9258 |

## 5.3 Evaluation metrics

In the fake review detection task, recall, precision, and the F-measure have been the most often used metrics. Since the number of reviews of both classes is equal, accuracy is also a popular metric for model evaluation. In order to evaluate the efficiency of the proposed model, we use the following metrics.

- Accuracy: full estimate of correctly classified instances and can be calculated by

$$Accuracy(Acc) = \frac{number\ of\ correct\ classifications}{total\ number\ of\ data\ samples} = \frac{TP + TN}{TP + TN + FP + FN} \qquad (1)$$

- Precision: describes the proportion of successfully predicted reviews to the total number of reviews for a given class and can be calculated by

$$Precision(P) = \frac{number\ of\ correct\ predictions\ of\ each\ class}{total\ number\ of\ predictions\ of\ each\ class} = \frac{TP}{TP + FP} \qquad (2)$$

- Recall shows the proportion of relevant reviews achieved from the total number of reviews and is calculated by

$$Recall\ (R) = \frac{number\ of\ correct\ predictions}{total\ number\ of\ predictions} = \frac{TP}{TP + FN} \qquad (3)$$

- F1 score shows the average of precision and recall and is calculated by

$$F-measure = \frac{2 \times recall \times precision}{recall + precision} \qquad (4)$$

## 5. RESULTS AND DISCUSSION

As shown in Table 3t the Acc, P, R, and F1-score for RoBERTa are 94.06%, 89.38%, 98.62%, 93.77%, respectively for Op-Spam dataset, 91.02%, 92.50%, 90.00%, 90.50% respectively for deception dataset. The experiments result of the XLNet model on the OpSpam dataset are as follows: Acc, P, R, and F1-score are 92.50%,86.25%, 98.57%, 92.00%, respectively, where the Acc, P, R, and F1-score on the deception dataset are 88.56%, 77.92%, 93.97%, 85.19%, respectively. On implementing ALBERT model on the OpSpam dataset are as follows: accuracy, precision, recall, and F1-score are 93.34%, 91.25%, 95.42%, 93.29%, respectively, where the Acc, P, R, and F1-score on the deception dataset are 88.03%, 77.08%, 93.43%, 84.87%, respectively. The experiments result of the proposed model on the OpSpam dataset are as follows: accuracy, precision, recall, and F1-score are 94.37%, 89.38%, 99.31%, 94.08%, respectively, where the Acc, P, R, and F1-score on the deception dataset are 92.07%, 84.17%, 96.65%, 89.98%, respectively. It is clear from the experiments that our proposed model achieved the best performance for both OpSpam and Deception datasets due to its efficiency in capturing the robust contextualised representation of each review and the most relevant features using different models and perspectives. Furthermore, our proposed model gains the full benefit of the most accurate transformer using the weighted average in addition to the appropriate support from other transformers.

Table 3. Classification reports for OpSpam and Deception datasets.

| | Datasets | | | | | | | |
|---|---|---|---|---|---|---|---|---|
| | OpSpam | | | | Deception | | | |
| Model | Acc | P | R | F1-score | Acc | P | R | F1-score |
| RoBERTa | 94.06% | 89.38% | 98.62% | 93.77% | 91.02% | **92.50%** | 90.00% | **90.50%** |
| XLNet | 92.50% | 86.25% | 98.57% | 92.00% | 88.56% | 77.92% | 93.97% | 85.19% |
| ALBERT | 93.43% | **91.25%** | 95.42% | 93.29% | 88.03% | 77.08% | 93.43% | 84.47% |
| **Proposed model** | **94.37%** | 89.38% | **99.31%** | **94.08%** | **92.07%** | 84.17% | **96.65%** | 89.98% |

## 6.1. Comparison with state-of-the-art models

To evaluate the effectiveness of the proposed model, our model is compared to the state of the art techniques, including SVM [5], SVM [19], SVM [15], SAGE [16], RCNN [38], GRNN–CNN [39], DRI-RCNN [27]. We use accuracy and F1-score according to the existing works [39], [27] as shown in Table 4.

- **SVM** [5]: A model of combining bigram and LIWC features using SVM as a classifier.
- **SVM** [19]: A model of a combination of four grams and LIWC features using SVM as a classifier.
- **SVM** [15]: A model of using unigram features with SVM as a classifier.
- **SAGE** [16]: The Sparse Additive Generative Model (SAGE) is a mix of topic modelling and a generalised additive model.
- **RCNN** [38] is a model of a combination of recurrent neural networks and convolutional neural networks.
- **GRNN–CNN** [39]: it is a hybrid fake reviews detection model. They combined a gated recurrent neural network (GRU) and a convolutional neural network.
- **DRI-RCNN** [27] is a recurrent convolutional deep neural networks model (DRI-RCNN) for detecting fake reviews based on word contexts.
- **BERT-Base Case** [6]: A BERT-trained model is used to pre-train a deep bidirectional representation of the text that is capable of handling unlabelled data by simultaneously focusing on right and left context in all layers.

From Table 4 results, it can be observed that the proposed model outperforms the current state-of-the-art methods on both the OpSpam and deception datasets. We can see that the ensemble of a pre-trained model achieved high accuracy with a small dataset compared to traditional machine learning and deep learning models. For example, our proposed model achieved over 90% accuracy with small training data, while deep learning could not reach 90%. Thus, we can observe that traditional and deep learning models require large datasets for training. However, conducting large datasets is not always possible. Therefore, the transformer model is a pragmatic option in the case of a small dataset.

Table 4. Comparison with the state-of-the-art methods.

| Machine Learning Models | Deception dataset | | Op-Spam dataset | |
|---|---|---|---|---|
| | Average Accuracy | F1-score | Average Accuracy | F1-score |
| SVM (bigram and LIWC features) [5] | 79.33% | 82.83% | 82.89% | 82.09% |
| SVM (unigram feature) [15] | 83.33% | 79.53% | 86.09% | 83.12% |
| SVM (four grams and LIWC features) [19] | 81.67% | 81.03% | 84.34% | 83.21% |
| SAGA [16] | 81.82% | 79.38% | 83.10% | 82.23% |
| RCNN [38] | 82.16% | 82.00% | 83.21% | 81.23% |
| GRNN-CNN [39] | 83.34% | 82.86% | 84.15% | 84.17% |
| DRI-RCNN [27] | 85.24% | 83.56% | 87.24% | 85.36% |
| BERT Base Case [6] | 86.20% | 85.50% | 90.31% | 89.56% |
| **Ensemble (Ours)** | **92.07%** | **89.98%** | **94.37%** | **94.08%** |

In the real world, detecting fake reviews is a constant challenge, and governments are working to resolve this issue in order to mitigate its negative impacts. Additionally, detecting fake news is a difficult assignment for a machine since it must understand the difference between "legitimate reviews" and "fake reviews." However, we may use a variety of features (e.g., reviews text and emotions) to produce an effective decision about detecting fake reviews. Thus, the more features we incorporate into our models during training, the more effective the model will be at detecting false reviews. In this paper, we employed a technique that is focused on deriving textual features from reviews text dove into the analysis of the differences between fake and real reviews content, indicating that the content of fake and real reviews is significantly different, and the style of fake reviews is more similar to satire than to actual reviews. As a result, contextual features play a critical part in determining the difference between fake and legitimate reviews, as demonstrated in our experiments. Also, according to the findings, the employment of pre-trained models and the ensemble approach significantly improved the results of detecting short fake reviews text.

There are two critical aspects that influence the performance of pre-trained language models: (1) the domain and size of the training datasets and (2) the model's architecture. RoBERTa derives from a diverse variety of data sources; for example, in addition to the standard BookCorpus and Wikipedia datasets, RoBERTa was trained using CC-News [40]. However, in the ensemble approach, the RoBERTa's performance was also improved by combining it with other transformers in the ensemble approach. This combination helps handle the dynamic change of fake review features and the variance in the provided context.

## 7. CONCLUSION AND FUTURE WORK

In this work, we investigated one significant research question for fake review detection which is **"Does the ensemble of transformer models perform better than state-of-the-art fake review detection methods?"** In order to answer this question, this paper investigated the performance of transformers in detecting fake reviews. We proposed a new model that combines three transformer based models, namely, RoBERTa, XLNet, and ALBERT and uses the weighted average for each classifier to obtain the best result. The proposed model on two semi-real datasets has shown 92.07% and 94.37 accuracies on OpSpam and deception datasets and outperformed the state-of-the-art methods, including traditional and deep learning models. Our proposed model performed significantly better than traditional and other deep learning models using small dataset sizes,demonstrating its efficiency. In future work, we will employ many textual analyses methods in natural language processing, such as named-entity recognition, to extract additional helpful information in addition to textual content embedding to detect fake reviews.

# REFERENCES

[1]   S. Saumya, J. P. Singh, A. M. Baabdullah, N. P. Rana, and Y. K. Dwivedi, "Ranking online consumer reviews," *Electronic commerce research and applications*, vol. 29, pp. 78-89, 2018.

[2]   J. P. Singh, S. Irani, N. P. Rana, Y. K. Dwivedi, S. Saumya, and P. K. Roy, "Predicting the "helpfulness" of online consumer reviews," *Journal of Business Research*, vol. 70, pp. 346-355, 2017.

[3]   S. Saini, S. Saumya, and J. P. Singh, "Sequential purchase recommendation system for e-commerce sites," in *IFIP International Conference on Computer Information Systems and Industrial Management*, 2017: Springer, pp. 366-375.

[4]   N. N. Ho-Dac, S. J. Carson, and W. L. Moore, "The effects of positive and negative online customer reviews: do brand strength and category maturity matter?," *Journal of Marketing*, vol. 77, no. 6, pp. 37-53, 2013.

[5]   M. Ott, Y. Choi, C. Cardie, and J. T. Hancock, "Finding deceptive opinion spam by any stretch of the imagination," in *Proceedings of the 49th annual meeting of the association for computational linguistics: Human language technologies-volume 1*, 2011: Association for Computational Linguistics, pp. 309-319.

[6]   R. Mohawesh *et al.*, "Fake Reviews Detection: A Survey," *IEEE Access*, vol. 9, pp. 65771-65802, 2021.

[7]   Y. Liu *et al.*, "Roberta: A robustly optimized bert pretraining approach," *arXiv preprint arXiv:1907.11692*, 2019.

[8]   Z. Yang, Z. Dai, Y. Yang, J. Carbonell, R. R. Salakhutdinov, and Q. V. Le, "Xlnet: Generalized autoregressive pretraining for language understanding," *Advances in neural information processing systems*, vol. 32, 2019.

[9]   Z. Lan, M. Chen, S. Goodman, K. Gimpel, P. Sharma, and R. Soricut, "Albert: A lite bert for self-supervised learning of language representations," *arXiv preprint arXiv:1909.11942*, 2019.

[10]  Y.-F. Huang and P.-H. Chen, "Fake news detection using an ensemble learning model based on self-adaptive harmony search algorithms," *Expert Systems with Applications*, vol. 159, p. 113584, 2020.

[11]  L. Shi, X. Ma, L. Xi, Q. Duan, and J. Zhao, "Rough set and ensemble learning based semi-supervised algorithm for text classification," *Expert Systems with Applications*, vol. 38, no. 5, pp. 6300-6306, 2011.

[12]  E. F. Cardoso, R. M. Silva, and T. A. Almeida, "Towards automatic filtering of fake reviews," *Neurocomputing*, vol. 309, pp. 106-116, 2018.

[13]  R. Mohawesh, S. Tran, R. Ollington, and S. Xu, "Analysis of Concept Drift in Fake Reviews Detection," *Expert Systems with Applications*, p. 114318, 2020.

[14]  J. W. Pennebaker, R. L. Boyd, K. Jordan, and K. Blackburn, "The development and psychometric properties of LIWC2015," 2015.

[15]  S. Feng, R. Banerjee, and Y. Choi, "Syntactic stylometry for deception detection," in *Proceedings of the 50th Annual Meeting of the Association for Computational Linguistics: Short Papers-Volume 2*, 2012: Association for Computational Linguistics, pp. 171-175.

[16]  J. Li, M. Ott, C. Cardie, and E. Hovy, "Towards a general rule for identifying deceptive opinion spam," in *Proceedings of the 52nd Annual Meeting of the Association for Computational Linguistics (Volume 1: Long Papers)*, 2014, pp. 1566-1576.

[17]  B. M. DePaulo, J. J. Lindsay, B. E. Malone, L. Muhlenbruck, K. Charlton, and H. Cooper, "Cues to deception," *Psychological bulletin*, vol. 129, no. 1, p. 74, 2003.

[18]  J. Hastie Trevor and J. Tibshirani Robert, "Generalized Additive Models. Vol. 43," ed: CRC Press, 1990.

[19]  L. Cagnina and P. Rosso, "Classification of deceptive opinions using a low dimensionality representation," in *Proceedings of the 6th workshop on computational approaches to subjectivity, sentiment and social media analysis*, 2015, pp. 58-66.

[20]  Q. Xu and H. Zhao, "Using deep linguistic features for finding deceptive opinion spam," in *Proceedings of COLING 2012: Posters*, 2012, pp. 1341-1350.

[21]  D. H. Fusilier, M. Montes-y-Gómez, P. Rosso, and R. G. Cabrera, "Detecting positive and negative deceptive opinions using PU-learning," *Information processing & management*, vol. 51, no. 4, pp. 433-443, 2015.

[22]  T. Young, D. Hazarika, S. Poria, and E. Cambria, "Recent trends in deep learning based natural language processing," *ieee Computational intelligenCe magazine*, vol. 13, no. 3, pp. 55-75, 2018.

[23]  M. Al-Hawawreh, E. Sitnikova, and F. den Hartog, "An efficient intrusion detection model for edge system in brownfield industrial Internet of Things," in *Proceedings of the 3rd International Conference on Big Data and Internet of Things*, 2019, pp. 83-87.

[24]  M. Al-Hawawreh and E. Sitnikova, "Industrial Internet of Things based ransomware detection using stacked variational neural network," in *Proceedings of the 3rd International Conference on Big Data and Internet of Things*, 2019, pp. 126-130.

[25]  Y. Ren and Y. Zhang, "Deceptive opinion spam detection using neural network," in *Proceedings of COLING 2016, the 26th International Conference on Computational Linguistics: Technical Papers*, 2016, pp. 140-150.

[26]  Y. Ren and D. Ji, "Neural networks for deceptive opinion spam detection: An empirical study," *Information Sciences*, vol. 385, pp. 213-224, 2017.

[27]  W. Zhang, Y. Du, T. Yoshida, and Q. Wang, "DRI-RCNN: An approach to deceptive review identification using recurrent convolutional neural network," *Information Processing & Management*, vol. 54, no. 4, pp. 576-592, 2018.

[28]  J. Devlin, M.-W. Chang, K. Lee, and K. Toutanova, "Bert: Pre-training of deep bidirectional transformers for language understanding," *arXiv preprint arXiv:1810.04805*, 2018.

[29]  C. Alippi, G. Boracchi, and M. Roveri, "A just-in-time adaptive classification system based on the intersection of confidence intervals rule," *Neural Networks*, vol. 24, no. 8, pp. 791-800, 2011.

[30]  Z. Dai *et al.*, "Attentive Language Models Beyond a Fixed-Length Context. arXiv 2019," *arXiv preprint arXiv:1901.02860*.

[31]  U. Naseem, I. Razzak, M. Khushi, P. W. Eklund, and J. Kim, "Covidsenti: A large-scale benchmark Twitter data set for COVID-19 sentiment analysis," *IEEE Transactions on Computational Social Systems*, 2021.

[32]  L. Prechelt, "Early stopping-but when?," in *Neural Networks: Tricks of the trade*: Springer, 1998, pp. 55-69.

[33]  L. Prechelt, "Automatic early stopping using cross validation: quantifying the criteria," *Neural Networks*, vol. 11, no. 4, pp. 761-767, 1998.

[34]  I. Loshchilov and F. Hutter, "Decoupled weight decay regularization," *arXiv preprint arXiv:1711.05101*, 2017.

[35]  L. Rosasco, E. De Vito, A. Caponnetto, M. Piana, and A. Verri, "Are loss functions all the same?," *Neural computation*, vol. 16, no. 5, pp. 1063-1076, 2004.

[36]  S. V. Halyal, "Running Google Colaboratory as a server–transferring dynamic data in and out of colabs," *International Journal of Education and Management Engineering*, vol. 9, no. 6, p. 35, 2019.

[37]  T. Wolf *et al.*, "Transformers: State-of-the-art natural language processing," in *Proceedings of the 2020 Conference on Empirical Methods in Natural Language Processing: System Demonstrations*, 2020, pp. 38-45.

[38]  S. Lai, L. Xu, K. Liu, and J. Zhao, "Recurrent convolutional neural networks for text classification," in *Twenty-ninth AAAI conference on artificial intelligence*, 2015.

[39]  Y. Ren and D. Ji, "Neural networks for deceptive opinion spam detection: an empirical study," *J Inf Sci*, vol. 385, 2017// 2017, doi: 10.1016/j.ins.2017.01.015.

[40]  S. Nagel, "Cc-news (2016)," ed.

## AUTHORS

**RAMI MOHAWESH** received the B.E and M.E in computer science and is now working towards a Ph.D degree from the University of Tasmania, Tasmania, Australia. In his PhD research, Rami is the first researcher who investigated the concept drift in fake review detection. He is a reviewer of high impact factor journals such as the Information Processing and Management Journal, Artificial Intelligence Review and Secure Computing. His research interests include Software Engineering, Cloud Computing, Natural Language Processing, Cybersecurity, and machine learning. His current work on Fake Review.

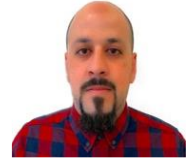

**SHUXIANG XU** is currently a lecturer and PhD student supervisor within the School of Information and Communication Technology, University of Tasmania, Tasmania, Australia. He has received a PhD in Computing from the University of Western Sydney, Australia, a Master of Applied Mathematics from Sichuan Normal University, China, and a Bachelor of Applied Mathematics from the University of Electronic Science and Technology of China, China. His research interests are Artificial Intelligence, Machine Learning, and Data Mining. Much of his work is focused on developing new Machine Learning algorithms and using them to solve problems in various application fields.

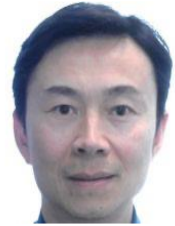

**MATTHEW SPRINGER** is a lecturer in the School of Technology, Environments and Design at the University of Tasmania. Dr Springer received his Information Systems PhD from the University of Tasmania in 2010. His major focus has been on improving teaching within the Discipline of Information and Communication Technology discipline but is also an active member of the Industry Transformation, and Games and Creative Technologies research groups.

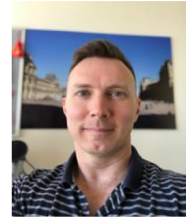

**MUNA AL-HAWAWREH** received the B.E. and M.E. degrees in computer science from Mutah University, Jordan. She is currently pursuing the Ph.D. degree with the University of New South Wales (UNSW), Canberra, Australia. She works as a Research Assistant at UNSW Canberra Cyber. In her Ph.D. degree, she developed the world's first ransomware framework targeting IIoT edge gateway in the critical infrastructure. Her research interests include cloud computing, industrial control systems, the Internet of Things, cybersecurity, and deep learning. She is a program committee member and a reviewer for several cybersecurity conferences. She was awarded the First Prize for high impact publications in the School of Engineering and Information Technology (SEIT), UNSW, in 2019, and the Dr. K. W. Wang Best Paper Award (2018–2020). She is a Reviewer of high-impact factor journals, such as the IEEE INTERNET OF THINGS JOURNAL, IEEE TRANSACTIONS ON INDUSTRIAL INFORMATICS, and IEEE TRANSACTIONS ON DEPENDABLE AND SECURE COMPUTING

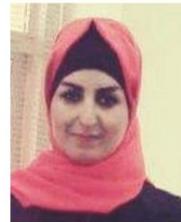

**SUMBAL MAQSOOD** received the BS. Hons Computer Science from Punjab University College of Information Technology (PUCIT) and MS Computer Science from GC University, Lahore Pakistan. She is currently second year Ph.D student at the University of Tasmania, Tasmania, Australia. She worked as an IT Officer in one of the biggest organisations of Pakistan. Her research interests include machine learning, natural language cybernetics, bio-technologies, data science and software engineering. She is currently working on biosignals analysis using deep learning.

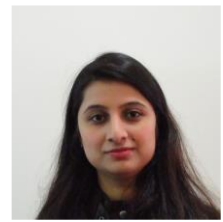